\documentclass[10pt,twocolumn,letterpaper]{article}

\usepackage{cvpr}
\usepackage{times}
\usepackage{epsfig}
\usepackage{graphicx}
\usepackage{amsmath}
\usepackage{amssymb}
\usepackage{subcaption}

\usepackage[breaklinks=true,bookmarks=false]{hyperref}

\cvprfinalcopy 

\def\cvprPaperID{****} 

\begin{document}

\title{Deep learning for 3D Object Detection and Tracking in Autonomous Driving: A Brief Survey}

\author{Yang Peng\\
Southern University of Science and Technology\\
{\tt\small 12032453@mail.sustech.edu.cn}
}

\maketitle

\begin{abstract}

Object detection and tracking are vital and fundamental tasks for autonomous driving, aiming at identifying and locating objects from those predefined categories in a scene. 
3D point cloud learning has been attracting more and more attention among all other forms of self-driving data. Currently, there are many deep learning methods for 3D object detection.
However, the tasks of object detection and tracking for point clouds still need intensive study due to the unique characteristics of point cloud data. 
To help get a good grasp of the present situation of this research, this paper shows recent advances in deep learning methods for 3D object detection and tracking.

\end{abstract}

\section{Introduction}

Object detection occupies an important position in the ﬁeld of computer vision all along. 
As the cornerstone of image understanding, object detection is widely applied in large numbers of areas such as autonomous driving and robot vision. 
Object detection enables an autonomous driving system to see those driving environments clearly and understand what are they just like human drivers. 

With the fast development of deep learning \cite{lan2022vision}, it becomes easier to learn complex, subtle and abstract features for a deep learning model without manual feature extraction like the traditional way. 
In view of the excellent ability of deep learning to process data, major progress has been made in the research of object detection. 

The research for 2D object detection has been well-developed \cite{lan2019evolving}. 
The accuracy and efﬁciency of algorithms proposed for solving 2D object detection problems have reached a high level and those methods play an important role in engineering practice. 
Traditional detection strategies based on sliding windows like \cite{DBLP:conf/cvpr/DalalT05, DBLP:journals/pami/FelzenszwalbGMR10} keep the mainstream for a long time before methods combining deep learning techniques. 
In this stage when deep learning hasn’t been applied to detection, the pipeline\cite{DBLP:journals/corr/abs-1908-03673} of object detection methods mentioned above usually can be categorized into three parts: i) proposal generation; ii) feature vector extraction; iii) region classification \cite{lan2022class}. 
The main task of proposal generation is to search the whole image to ﬁnd those positions that might contain expected objects. 
These positions are called regions of interest (ROI). 
The sliding window technique is an intuitive idea proposed to scan through the input image and ﬁnd ROI. In the second stage, algorithms will extract a constant-length feature vector from each target position of the image. 
Histogram of Gradients\cite{DBLP:conf/cvpr/DalalT05} is one of the most popular feature extraction methods proposed by Navneet Dalal and Bill Triggs. 
In general, linear support vector machines are used together with HOG, achieving region classiﬁcation.

With these traditional detectors, much success has been achieved in object detection. Nevertheless, there are still some limitations here. 
The huge search space and expensive computations of methods based on sliding windows encourage better solutions. 
Moreover, one shortage which should not be overlooked is that the procedure of design and optimize a detector is separate \cite{lan2022time}. 
This may lead to local optimal solution for the system.

\begin{figure*}[!ht] \centering \small
    \includegraphics[width=0.95\linewidth]{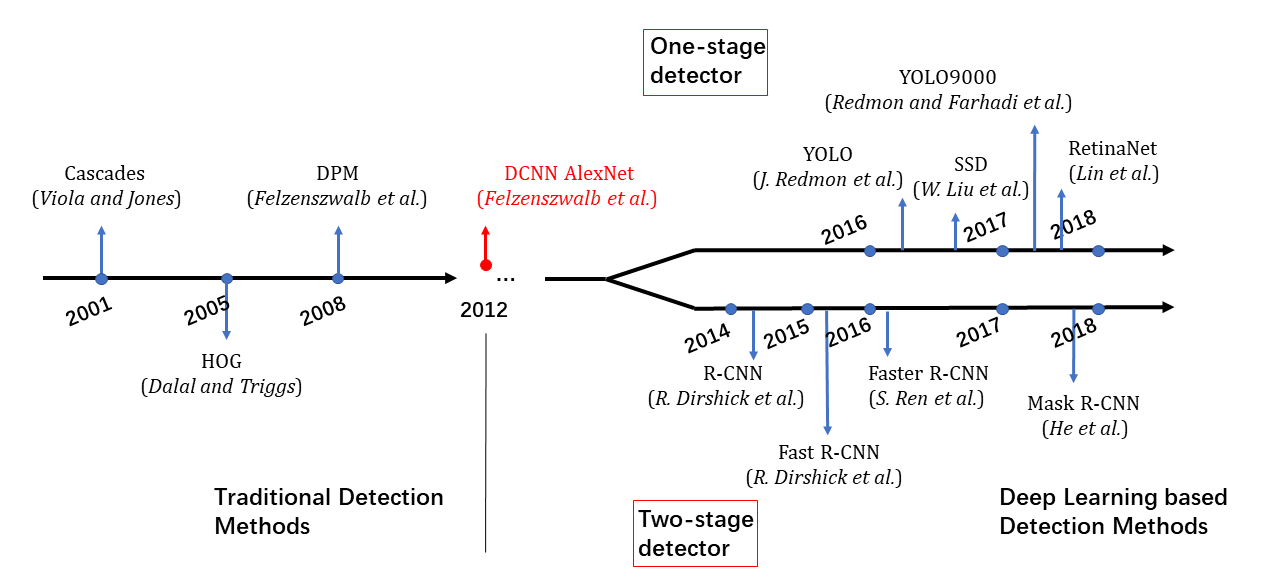}
   \caption{Major developments and milestones of object detection techniques. In 2012 there occurred a turning point drawing a line between traditional and deep learning-based object detection methods.
}
\label{fig:timeline}
\end{figure*}

Since deep learning emerged in this ﬁeld, it has become more capable of detecting objects. 
These object detection systems using deep learning techniques can handle questions with more facility and rapidity than in the past. 
We observe a turning point in 2012 with the development of DCNN for image classiﬁcation by Krizhevsky et al.\cite{DBLP:conf/nips/KrizhevskySH12}. Since this time, there has been an incessant springing up of approaches based on deep learning. 
Currently, the mainstream of object detection algorithms can be divided into two types: i)two-stage methods, such as Region-based CNN (R-CNN) \cite{DBLP:conf/cvpr/GirshickDDM14} and its variants\cite{DBLP:conf/iccv/Girshick15,DBLP:conf/iccv/HeGDG17,DBLP:conf/nips/RenHGS15}; ii)one-stage methods such as You Only Look Once (YOLO)\cite{DBLP:conf/cvpr/RedmonDGF16} and its variants\cite{DBLP:conf/cvpr/RedmonF17}. Two-stage methods are one kind of region proposal-based methods, which propose some possible regions that contain objects ﬁrstly and extract feature vectors from these regions respectively. One-stage methods predict the categories of objects on the location of the feature maps directly to the contrary, discarding the region classiﬁcation steps. 
The different kinds of methods have their own strong points. 
Fast-RCNN\cite{DBLP:conf/iccv/Girshick15} cut down the processing time of the previous network. However, a bottleneck bothers people a lot because the computation of this network is quite expensive. 
Until the appearance of Faster-RCNN\cite{DBLP:conf/nips/RenHGS15} this problem gets solved. 2D detectors are not the key point in this paper. 
For the integrity of the whole review and introduction of 3D object detection methods, we mention about those methods. 
Here we present the main progress and milestones of 2D object detection techniques in Figure\ref{fig:timeline}. 

However, with new application scenarios such as robot vision and autonomous driving proposed \cite{lan2018real}, the implementation of 2D object detection is far from enough. 
The process of images captured by cameras is projecting 3D space into 2D view, which causes the loss of 3D spatial information and is unable to satisfy people. 
More 3D spatial information needs to be considered. 
With the rapid development of many 3D techniques, large amounts of 3D sensors such as LiDARs, 3D scanners and RGB-D cameras are becoming increasingly available and affordable \cite{xu2019online}. 
In this review, we mainly analyse point clouds obtained by using LiDARs. As a common format, each point provides us with useful geometric position information and some may contain RGB information. 
There is an urgent demand for 3D object detection and tracking when it comes to autonomous driving. In the research of autonomous driving, accurate environmental perception and precise location are the keys for an autonomous driving system to achieve reliable navigation, information decisions and safe driving in complex and dynamic environments. 
Compared to images whose data quality is influenced by illumination, point cloud is robust for different lighting conditions [19]. As a result, point cloud data is widely used in this ﬁeld. 

\begin{figure}
\centering
\includegraphics[width=0.95\linewidth]{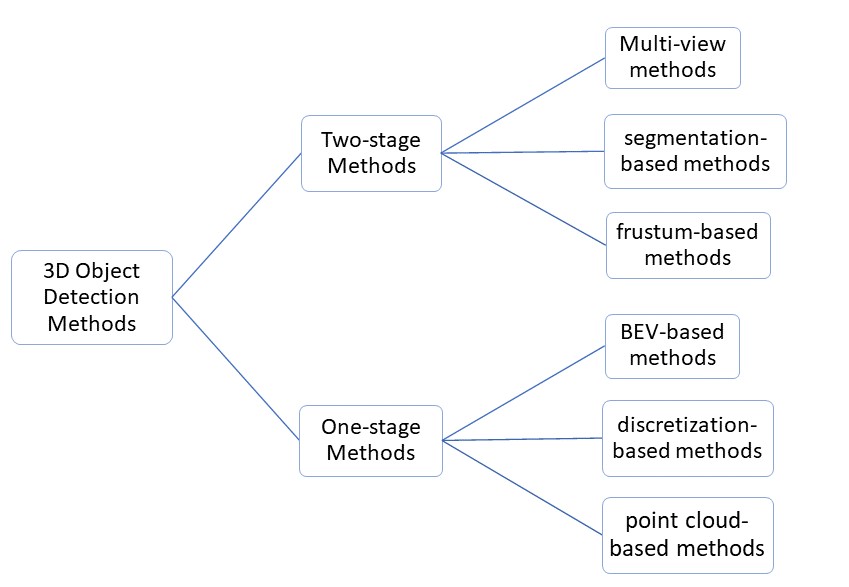}
\caption{Summary of major deep learning-based methods of 3D object detection.
}
\label{fig:3d}
\end{figure}

On the basis of 2D object detection, changes have taken place in the questions and requirements of object detection. 
For the sparsity and the irregularity of point clouds, it is impossible to apply 2D object detection methods to 3D point clouds. 
As a result, 2D object detection methods need to be changed to make sure those methods can be extended to 3D cases. Similar to the object detection approaches applied in images, these 3D object detection methods still can be divided into two categories: two-stage methods and one-stage methods. Figure\ref{fig:3d} presents a simple classiﬁcation of recent popular 3D object detection methods based on deep learning techniques. 
More details of these 3D detection methods can be found later Section 3 of this paper. 

For this paper, we mainly focus on these state-of-the-art methods of 3D object detection on the basis of point clouds. 
The remaining parts of this review are organized as follows. 
Related backgrounds, including the problems identiﬁcation, the questions description, the key challenges and the motivation of this review are stated in Section \ref{Background}. 
In the third part of this paper, we make a detailed interpretation of those current approaches for 3D object detection, including their advantages and disadvantages.
Section 4 makes a summary of recent developments in deep learning methods for point clouds. 
Furthermore, we point out some difﬁcult problems still unsolved on the basis of existing methods and provide several possible research directions that might make sense.

\begin{figure*}[!ht] \centering \small
    \includegraphics[width=0.95\linewidth]{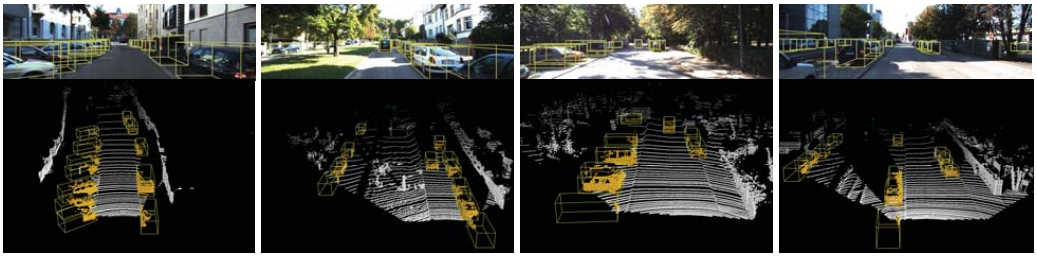}
   \caption{ Bounding boxes.
}
\label{fig:bbx}
\end{figure*}

\section{Background}

For the purpose of helping understand how exactly modern deep learning techniques can be applied to solve 3D object detection problems, this section describes some simple conceptions of the research object and the data.

\subsection{The Problem}

The problem of 3D object detection, which is similar to 2D detection, can be formulated as follows. Given some point cloud data, a 3D detector needs to determine whether or not there are instances from those predeﬁned categories and if yes, locate where they are. One thing we should ﬁrst consider is these predeﬁned categories, which are problem-dependent. Considering the case of autonomous driving, the research community is more interested in those trafﬁc participants (e.g., bicycles, cars and pedestrians). In general, unstructured scenes such as sky and clouds are often not our research objects. 

In general, a 3D bounding box is used to indicate the location of those 3D objects. It is a rectangular cuboid placed in 3D spaces and usually has three types of representation methods, such as axis-aligned 3D centre offset method\cite{DBLP:conf/cvpr/SongX16}, 8-corners method\cite{DBLP:conf/cvpr/ChenMWLX17} and 4-corner-2-height method\cite{DBLP:conf/iros/KuMLHW18}. In general, the ﬁnal result of 3D object detection is using 3D bounding boxes to earmark 3D objects, as shown in Figure\ref{fig:bbx}.

\subsection{Main Challenges}

Generic object detection methods aim at recognizing and localizing interested object categories. In most computer vision problems, people care about accuracy and efﬁciency mostly. Object detection is no exception. 

Those point cloud data acquired from LiDARs, depth cameras and binocular cameras can be used in 3D object detection. However, with the increase in the distance between objects and cameras, there would be a sudden drop in the density of the point clouds, causing a huge change in the density. What’s more, a few portions of the objects might be invisible owing to the occlusion, which causes a problem that there exists a large lag in the distribution of the point clouds of the same object.
To sum up, the representations of point clouds are quite different and hard for a detector to make fairly accurate detections. 

Another point we should notice is the sparsity and the irregularity of point clouds. 
The order of those points of the same object is greatly influenced by different acquisition devices and different coordinate systems. Those irregular cloud points make it really hard for an end-to-end model to handle. 
In addition, compared with the large scale of scenes, the coverage of LiDAR sampling points has strong sparsity. 
It is well-known that with the rapid development of the artiﬁcial intelligence, deep neural network is widely used in most tasks of autonomous driving because of their high accuracy and strong robustness. The performance of deep neural networks \cite{lan2021learning,lan2021learning2} used in the ﬁeld of 2D object detection is much better than other types of algorithms. Nevertheless, we have just mentioned the characteristics of point clouds, which result in the loss of efﬁcacy of those deep learning methods. This is why the movement of research in deep learning-based 3D object detection is slow. As a result, in the stage of data preprocessing, how to represent sparse point cloud data for better use deserves to be investigated thoroughly. 

Despite the useful information such as depth and spatial information that point clouds have, it seems that utilizing image data at the same time performs better sometimes. 
Therefore a number of methods combining LiDAR point clouds and images are well developed. 
In this paper, we mainly talk about deep learning methods for point clouds. 
For the completeness of this paper, some fusion methods which use 2D images as well will also be included. 

\subsection{Motivation}

Related methods mentioned in this paper are organized based on their different algorithm execution processes. 
In other words, we mainly focus on whether a detector ﬁrst generates proposals. 
This paper intends to investigate state-of-the-art research about 3D object detection for point clouds and categorize those methods, presenting a comprehensive summary of recent advances based on deep learning technologies for point clouds. 
It also covers the comparative advantages and disadvantages of different methods, by observing those questions still unsolved to inspire more future possible research directions.

\section{Methods}



\subsection{Two-Stage Methods}

Two-stage detectors ﬁrst detect a number of possible regions also called proposals which contain objects, then make predictions for extracted features. According to the introduction of \cite{DBLP:journals/pami/GuoWHLLB21}, we further categorize those two-stage methods into three kinds: i) multi-view-based; ii) segmentation-based; iii) frustum-based methods. 

\begin{figure}
\centering

\begin{subfigure}[t]{\linewidth}
\includegraphics[width=0.95\linewidth]{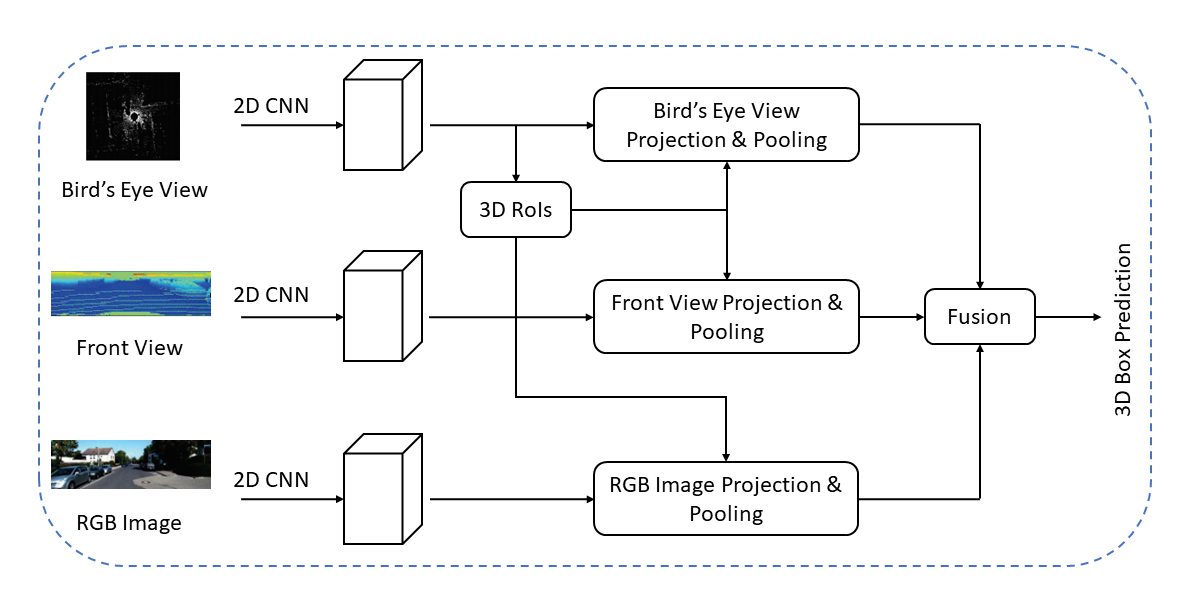}
\caption{MV3D}
\label{fig:mv3d}
\end{subfigure}

\begin{subfigure}{\linewidth}
\includegraphics[width=0.95\linewidth]{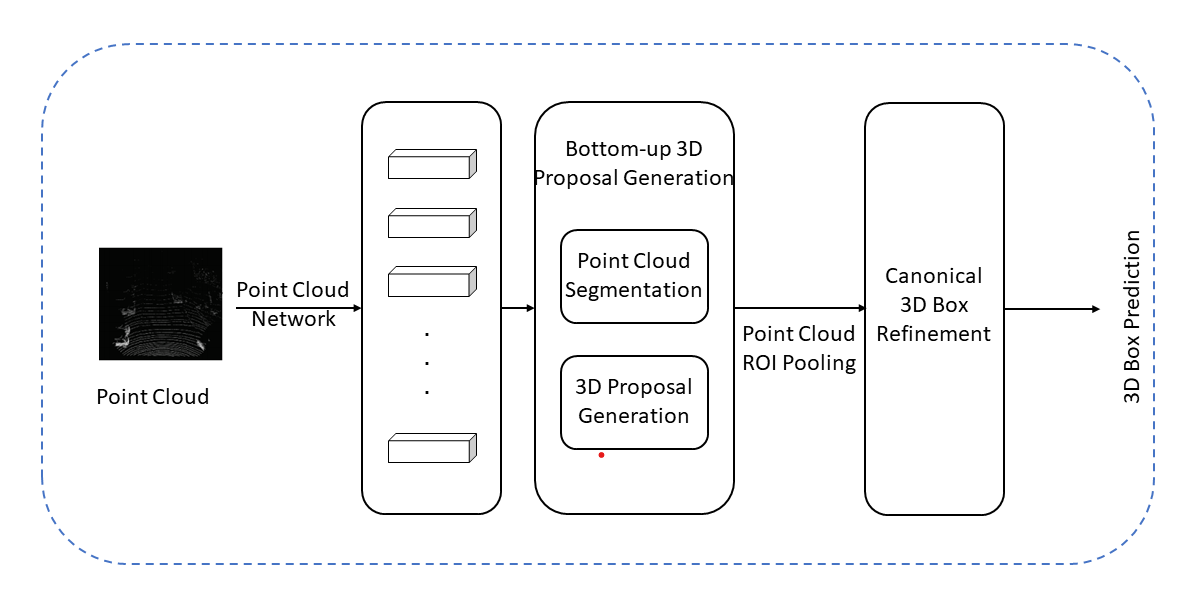}
\caption{PointRCNN}
\label{fig:prcnn}
\end{subfigure}

\begin{subfigure}{\linewidth}
\includegraphics[width=0.95\linewidth]{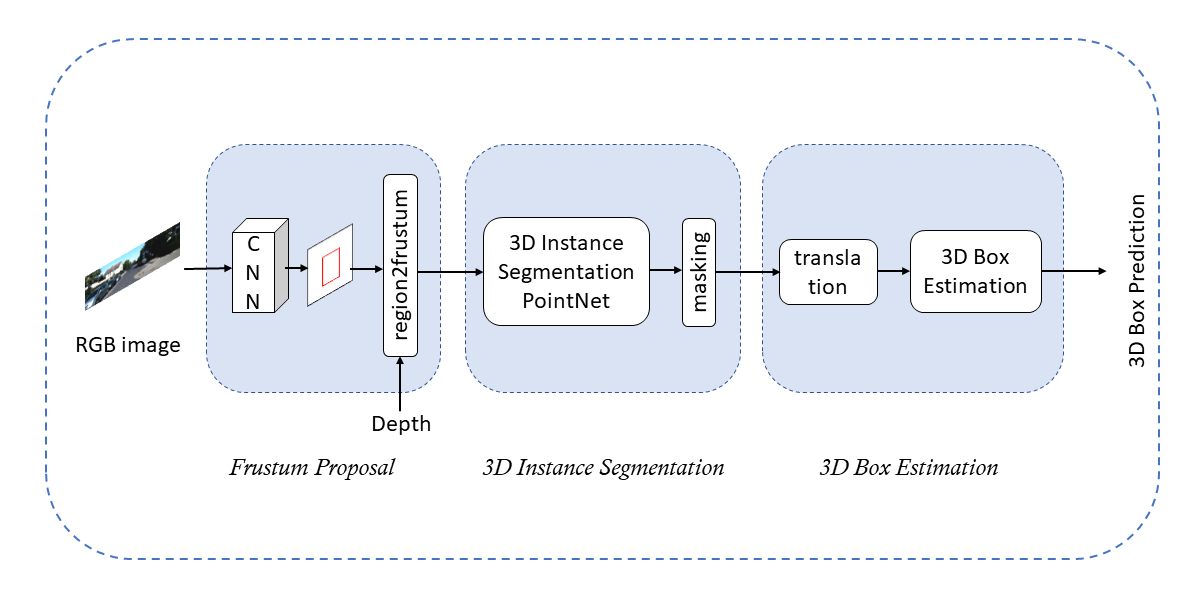}
\caption{Frustum PointNet}
\label{fig:f-pointnet}
\end{subfigure}

\caption{Typical networks for 3D object detection methods.
}
\end{figure}

\textbf{Multi-View Methods.}  It is necessary to realize that texture information, which is important for class discrimination is not included in point clouds. By contrast, monocular images cannot provide us with depth information for accurate 3D localization and size estimation. So multi-view methods try to use different modalities to improve the performance. A deep fusion combines region-wise features from multiple views (e.g., bird’s eye view (BEV), LiDAR front view (FV) and images) and gets oriented 3D boxes, as shown in Figure \ref{fig:mv3d}. 

One remarkable contribution made by Chen et al. \cite{DBLP:conf/cvpr/ChenMWLX17} is their multi-view 3D object detection network (MV3D). In terms of the expensive computations of existing LiDAR-based 3D methods, the disadvantage of current image-based methods and the limitation of existing multi-modal fusion methods, this work tries to propose a method to overcome those disadvantages preventing the development of 3D object detection. The MV3D is composed of two subnetworks: one is a 3D proposal network to generate 3D proposals and another subnetwork is a deep fusion network, whose main function is trying to fuse multi-view feature proposals. They use 3D proposals and project them to three views. Theoretically, it is of great useful that 3D proposals can be projected to any view in 3D space. In their experiments, LiDAR-based methods achieve an average precision of 87.65\% at an Intersection-over-Union (IOU) of 0.5 on the KITTI validation set, which obtains 30\% higher over VeloFCN \cite{DBLP:conf/rss/LiZX16}. 
Besides, the authors illustrate the construction process of BEV and state three advantages of BEV, which casts light on their search for a bird’s eye view and has a certain reference value for future research. 
Nevertheless, this model is not perfect because the whole procedure for detection is too slow to have practical applications. 
This region proposal network(RPN) is not suitable for small object instances in BEV. 
Small objects occupy a fraction of a pixel in a feature map, causing insufﬁcient data to extract features. Subsequently, a number of endeavours have been made to improve the MV3D model. 

Several attempts are made on the efﬁciency of information fusion by applying different modalities. Ku \textit{et al.} \cite{DBLP:conf/iros/KuMLHW18} proposed an aggregation view object detection network (AVOD), which uses LIDAR point clouds and RGB images as well. Unlike MV3D it extends ROI feature fusion to the stage of proposal generation. 
Their RPN has a novel architecture having the ability to ﬁnish multi-modal feature fusion on high-resolution feature maps, thus it is convenient to generate accurate region proposals for tiny objects in scenes. 
Furthermore, they test the performance of AVOD on the KITTI Object Detection Benchmark \cite{DBLP:conf/cvpr/GeigerLU12} and the result shows that AVOD can run in real-time and with a low memory expense. 
However, ROI feature fusion is conﬁned to high-level feature maps. 
Furthermore, only those features extracted from selected object regions will be fused. 
ContFuse \cite{DBLP:conf/eccv/LiangYWU18} was developed to overcome those drawbacks. They exploited continuous convolutions to fuse feature maps at different resolutions. With the projection of the LIDAR points image and BEV spaces can correspond with each other. In other words, in the BEV spaces corresponding image features for each point can be extracted and projecting image features into the BEV plane can obtain dense BEV feature maps. However, extremely sparse point clouds set limits on this kind of fusion. 
Liang \textit{et al.} \cite{DBLP:conf/cvpr/LiangYCHU19} proposed an object detection network for multiple tasks (e.g., 3D object detection, ground estimation and depth completion) with multi-sensors. This paper utilizes the strong points from both point-wise and ROI-wise feature fusion. Speciﬁcally, the implementation of multiple tasks helps whole network learn better representations. 
Consequently, the KITTI and TOR4D datasets are used for validation of this approach, which proves to achieve outstanding improvement on detection problems and outperforms state-of-the-art approaches in the past. 

Another direction for improvement is to explore how to extract robust representations of the inputs. 
A novel Spatial-Channel Attention Network (SCANet) \cite{DBLP:conf/icassp/LuCZZMZ19} was proposed aiming at achieving high accuracy 3D object detection. 
In this paper, they raise a brand-new Spatial-Channel Attention (SCA) module and an Extension Spatial Upsample (ESU) module for 3D region proposal use. 
The former module can focus on global and multi-scale contexts in a scene and can capture those discriminative features. 
The latter module combines different scale low-level features and produces reliable 3D region proposals. 
In addition, a way to fuse those features better is to apply a new multi-level fusion scheme allowing more interactions between them. Finally, the experimental results present that at a speed of 11.1 FPS their method is 5 times faster than MV3D. 

\textbf{Segmentation-based Methods.} This kind of method usually has preliminary processing of semantic segmentation \cite{lan2022semantic}, which means that using semantic segmentation techniques to take out those background points can help generate high-quality proposals from foreground points. 
One typical segmentation-based network is shown in Figure \ref{fig:prcnn}. 
In addition, compared with those multi-view methods mentioned above, segmentation-based methods obtain higher recall scores and can be applied in complex scenes with a large number of occlusions and crowded object swell. 

The first method to be introduced is proposed in \cite{DBLP:journals/corr/abs-1812-05276}. 
This method named IPOD ﬁrst completes semantic segmentation on the images and generates point-based proposals. 
Generating proposals at those positive points remains high ﬁdelity as well. Besides, some possible questions such as proposal redundancy and ambiguity have also been considered. This paper provided a new criterion named PointsIoU to address those problems. Experimental results surely show that this model is better than many 3D detection methods, especially for those scenes with high occlusion.

Another classical segmentation-based network in our review to talk about is PointRCNN \cite{DBLP:conf/cvpr/ShiWL19} proposed by Shi \textit{et al.} This network generates 3D proposals via segmentation techniques in the ﬁrst stage and in stage two those proposals will be reﬁned to get ﬁnal detection results. Unlike IPOD, PointRCNN directly segments the point cloud to generate high-quality proposals rather than applying 2D object segmentation. An important module of this network is Bin-based 3D bounding box generation and those boxes are regressed from the foreground points. Instead of using L1 or L2 loss for regression, this module adopts a bin-based method. That is to say, they ﬁrst split each foreground point into different bins, then regress boxes among each bin. This work achieves an RPN in 3D space. Drawing lessons from the RPN stage of PointRCNN, Jesus \textit{et al.} proposed a graph-based 3D detection pipeline named PointRGCN \cite{DBLP:journals/corr/abs-1911-12236}, which takes advantage of advances in GCNs, including two subnetworks R-GCN and C-GCN. R-GCN is a residual GCN that achieves pre-proposal feature aggregation by using all points in a proposal. C-GCN is a contextual GCN whose main function is reﬁning proposals via the shared contextual information between different proposals. Sourabh \textit{et al.} proposed PointPainting \cite{DBLP:conf/cvpr/VoraLHB20} which works by projecting lidar points into the output of an image-based semantic segmentation network and appending the class scores to each point. Those appended points can be fed to any existing lidar-only detector such as PointRCNN we have mentioned above. Their work can ﬁll the gap that comprehensive information provided by different sensors is beneficial for fusion-based methods, but experimental results on the main benchmark datasets show that lidar-only methods outperform fusion-based methods in most cases. 

\textbf{Frustum-based Methods.} These methods leverage both mature 2D object detectors and advanced 3D deep learning for object localization. 
They generate 2D object region proposals ﬁrst and then generate 3D frustum proposals via lifting a 2D bounding box to a frustum that includes a 3D search space for the object. 
One speciﬁc process of the frustum-based method can be seen in Figure \ref{fig:f-pointnet}. Certainly, it is of great significance to know that although with high efﬁciency to propose possible regions of 3D objects, the step-by-step pipeline results in extreme dependency on 2D image detectors. 

Qi \textit{et al.} did pioneering work in this direction. They proposed a novel framework named Frustum PointNets \cite{DBLP:conf/cvpr/QiLWSG18}, based on RGB-D data for 3D object detection. In their work, the model ﬁrst feeds RGB images to the convolutional neural network to obtain 2D proposals and then combines depth information to project region to frustum. This is the process of getting frustum proposals. For those points contained in the frustum, 3D instance segmentation would be executed. Based on the results of segmentation, a lightweight regression PointNet attempts to adjust those points via translation so that their centroid is close to amodal box centre. 
Finally, a 3D box estimation network estimates those 3D amodal bounding boxes. It is surprising that F-PointNets have the ability to predict correctly posed amodal 3D boxes with a few points. However, there still exists some problems such as failure to work in the case of multiple instances from the same category. Following the work of F-PointNets, Zhao \textit{et al.} presented a new network architecture called SIFRNet \cite{DBLP:conf/aaai/ZhaoLHH19} relying on front view images and frustum point clouds to predict 3D detection results. The whole network mainly consists of three parts: i)3D instance segmentation network (Point-UNet); ii)T-Net; iii)3D box estimation network (Point-SENet). They contribute to the improvement of the performance in 3D segmentation and the efﬁciency of 3D bounding box prediction. PointSIFT \cite{DBLP:journals/corr/abs-1807-00652} module is integrated into their network, capturing orientation information of point clouds and achieving strong robustness to shape scaling. A series of experiments show that this method achieves better performance on the KITTI dataset and SUN-RGBD dataset \cite{DBLP:conf/cvpr/SongLX15} when compared to F-PointNets. 

A generic 3D object detection method called PointFusion \cite{DBLP:conf/cvpr/XuAJ18} is presented by Xu \textit{et al.} To handle the challenge of the combination of various RGB images and point cloud data, previous methods usually transform the form of point cloud data such as representing point cloud data by 2D image or voxel. It is convenient but runs into the problem of losing some information contained in point clouds. Instead, this method directly processes images and 3D point clouds via ResNet \cite{DBLP:conf/cvpr/HeZRS16} and PointNet \cite{DBLP:conf/cvpr/QiSMG17}. 
The obtained 2D image region and its related frustum points are used to precisely regress 3D boxes. They present a global fusion network to directly get the 3D box corner locations. At the same time, a new dense fusion network for the purpose of predicting spatial offset and choosing the ﬁnal prediction results with the highest score. 

It is noteworthy that Wang \textit{et al.} put forward a novel approach called Frustum ConvNet \cite{DBLP:journals/corr/abs-1903-01864}, which ﬁrst generates a strand of frustums for each proposal and utilizes obtained frustums to group those points. F-PointNet we have mentioned above also works directly on raw point cloud, but it has not been designed to be a end-to-end pipeline, owing to its T-Net alignment. This method takes this factor into account and is designed to combine several beneﬁts from previous works. It has been proved that this novel method F-ConvNet for amodal 3D object detection with end-to-end style achieves the state-of-the-art performance on the KITTI dataset among 2D detectors and is helpful for a large number of applications such as autonomous driving. 

Two-stage methods obtain several advantages including high accuracy for object detection. However, because of the process of generating regions containing pre-deﬁned objects, the speed of this object detection procedure will be decreased. 

\begin{figure*}
\begin{center}
\includegraphics[width=0.95\linewidth]{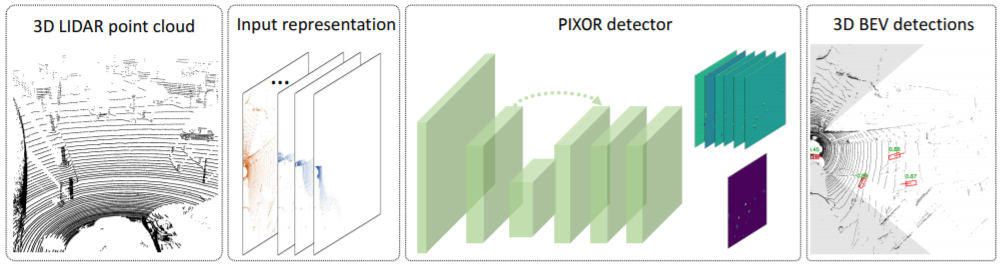}
\end{center}
   \caption{A comprehensive view of the proposed 3D object detector from Bird’s Eye View (BEV) based on point clouds.
}
\label{fig:pixor}
\end{figure*}

\subsection{One-Stage Methods}

In our discussion, one stage is equal to a single shot in the meaning of predicting class probabilities and regressing bounding boxes directly. 
These methods are free of region proposal network (RPN) and post-processing. 
Consequently, they have higher speed than two-stage approaches and are applied to real-time systems mostly. 
We categorize one-stage methods to the form of input data, including BEV-based methods, discretization-based methods and point cloud-based methods. 

\textbf{BEV-based Methods.} Looking into the name of this kind of method, it is obvious that BEV-based methods use BEV representation as input. As mentioned in \cite{DBLP:conf/cvpr/ChenMWLX17}, the BEV map has several strong points in 3D object detection compared to the front view or image plane. 
First, in the BEV map objects are of the same physical sizes as the original sizes, thus decreasing the size error that the front view/image plane has. 
Second, it is almost impossible to run into occlusion problems because objects in the BEV are located in different places. 
Last, for autonomous driving applications, those trafﬁc participants usually lie on the ground plane and in the vertical direction there exists a small variance. 
So the BEV location is of great significance when getting 3D bounding boxes. 

Yang \textit{et al.} proposed a proposal-free, one-stage method named PIXOR \cite{DBLP:conf/cvpr/YangLU18}, which represents raw 3D data from the Bird’s Eye View (BEV). We can see the network architecture of PIXOR from Figure \ref{fig:pixor}. Here the BEV representation is chosen because of its friendly computation. In the stage of elaborating input representation, they discretize 3D points contained in those interested scenes with a resolution of $d_L$ × $d_W$ × $d_H$ each cell ﬁrst and then encode the value of each cell to get occupancy tensor. 
Finally, they get a combination of the 2D reﬂectance image and the 3D occupancy tensor. When designing network architecture, a fully convolutional network is used for dense 3D object detection. A backbone network and a header network are composed of PIXOR. 
The former extracts ordinary representation of the input and the latter is used for doing speciﬁc task prediction. 
Based on experimental results, the conclusion has been made that PIXOR outperforms most one-stage methods in terms of Average Precision (AP) while running at 10 FPS.

Later on, Yang \textit{et al.} \cite{DBLP:conf/corl/YangLU18} showed us that High-Deﬁnition (HD) maps possess strong priors, which is beneﬁcial for the improvement of the performance and robustness of 3D object detection. 
To reach their goal, they developed a single-stage detector that works in the Bird’s Eye View (BEV) and fuses LiDAR information. 
Speciﬁcally, they get the coordinates of ground points from the HD map and then the absolute distance in the BEV representation would be replaced by the distance relative to the ground for the purpose of mending translation variance. 
Considering the situation that HD maps are not available everywhere, they put forward a solution that is practicable to utilize an online map prediction module and estimate map priors on the basis of LiDAR point cloud data. 
In their experiments, the baseline is a PIXOR++ detector without a map. 
The results illustrate that this HD map-aware model outperforms its baseline on the TOR4D and KITTI evidently. 
Nevertheless, a problem still exists for the poor generalization performance to point cloud data with diverse densities. Intending to address this problem, a new network architecture was proposed by Beltran \textit{et al.} named BirdNet \cite{DBLP:conf/itsc/BeltranGMC0E18}. 
They developed a new encoding method for BEV, which is invariant to distance and differences on LiDAR devices. The proposed density normalization method enables training models on those popular high-resolution lase datasets. 

\textbf{Discretization-based Methods.} It is easy to understand this kind of method via its name. These methods usually convert raw point cloud data into a regular discrete format (e.g., 2D map), and then use deep neural networks to predict the category probabilities and 3D boxes of objects. 

\begin{figure*}[!ht] \centering \small
\includegraphics[width=0.95\linewidth]{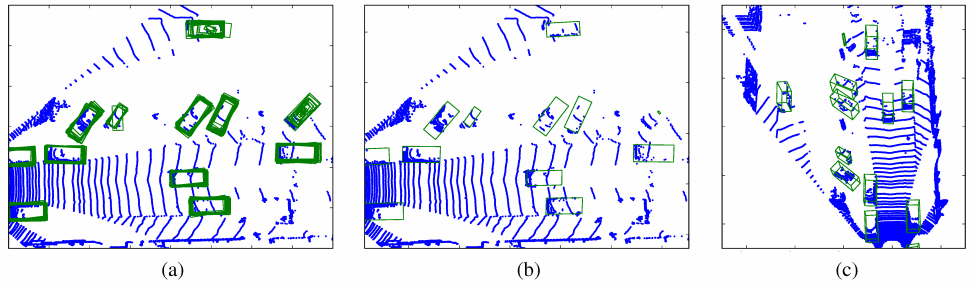}
   \caption{Direct outcomes of the 3D FCN detection process. (a) Aggregating bounding boxes in possession of high objectness conﬁdence and plotting them as green boxes. Plotting bounding box predictions coming from with green boxes. (b) Plotting those bounding boxes with blue original point clouds after clustering. (c) Detection in 3D since (a) and (b) are visualized in the bird’s eye view.}
\label{fig:fcn}
\end{figure*}

The first method to use a FCN in the purpose of 3D object detection was proposed by Li \textit{et al.}. Owing to the quick development of convolutional network techniques, they raised a new idea to transplant the FCN technique to object detection on 3D data \cite{DBLP:conf/rss/LiZX16}. 
In this method, point cloud data are converted into a 2D point map and the FCN is used for predicting the conﬁdences of those objects and the bounding boxes at the same time. 
This is the first attempt to introduce the FCN techniques into the object detection on range scan data, generating an order and end-to-end framework for detection. 
However, this paper only analyzes the method for 3D range scan from Velogyne 64E. As a result, in their later work, they discretized the point cloud into a 4D tensor with dimensions of length, width, height and 
channels \cite{DBLP:conf/iros/Li17}. In this work, they also extended the former 2D detection method based on FCN to 3D domain for 3D object detection. We can observe the detection results in Figure \ref{fig:fcn}. Based on their experimental results, they made a comparison with FCN for 2D detection work, observing a gain of over 20\% in accuracy. 

Although 3D FCN-based method makes progress compared to the previous work, there are still some possible questions here. 
Recalling the process, it is easy to ﬁnd that this method has a huge expensive computation because of 3D convolutions and the sparsity of the data. Engelcke \textit{et al.} put forward a computationally efﬁcient approach named Vote3Deep to do object detection for 3D point clouds \cite{DBLP:conf/icra/EngelckeRWTP17}. 
Vote3Deep is a feature-centric voting framework for the purpose of improving the efﬁciency of computation.
For one thing, building an efﬁcient convolutional layer via leveraging a voting framework is of great use.
For another, to take advantage of sparse convolutional layers of the whole CNN stack they utilize modiﬁed linear units and sparsity penalty. Li \textit{et al.} proposed a 3D backbone network learning 3D features from most raw data \cite{2019Three}. 
The network is constructed to address the problem of the lack of powerful 3D feature extraction methods. As a result, this method obtains rich 3D features and would not introduce a huge computational burden. 

\begin{figure}
\begin{center}
\includegraphics[width=0.95\linewidth]{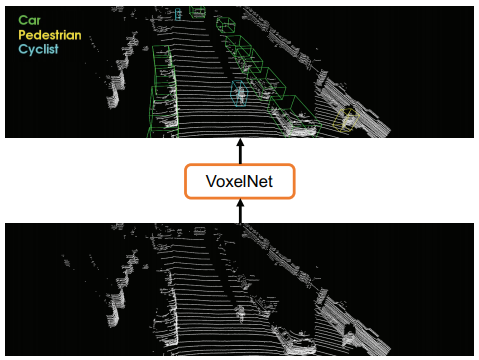}
\end{center}

   \caption{VoxelNet directly works on the raw point cloud without feature engineering and predicts the 3D detection results using a single end-to-end trainable network.
}
\label{fig:voxelnet}
\end{figure}

Zhou \textit{et al.} proposed an end-to-end network named VoxelNet for 3D object detection \cite{DBLP:conf/cvpr/ZhouT18}. Figure \ref{fig:voxelnet} shows how VoxelNet works. They projected point clouds into equally spaced voxels and transformed a number of points in each voxel into a centralized feature representation via the brand-proposed voxel feature encoding (VFE) layer. The method using skill to represent point clouds with voxels has great performance. Voxel-based methods sometimes are designed with the intention of enhancing the retention of information at the stage of processing point cloud data. However, the speed of the model process is still very slow because of the sparsity of those voxels and 3D convolutional computations. Following Zhou’s work, Yan \textit{et al.} did research on how to increase the speed of training and prediction and proposed an improved sparse convolution method \cite{DBLP:journals/sensors/YanML18}. What’s more, to handle the question that it is ambiguous to solve the sine function between 0 and $\pi$, they proposed a sine error angle loss. Another attempt has been made by Sindagi \textit{et al.} about a small change on VoxelNet is fusing image and point cloud features at a different stage \cite{2019arXiv190401649S}. Specifically, PointFusion and VoxelFusion are two unique techniques mentioned in \cite{2019arXiv190401649S}. The former works in the early stage where point cloud data can be projected to the image plane. The latter is used to project those 3D voxels to the image. Compared to VoxelNet, this network with new techniques has the ability to exploit multi-modal information and can decrease the false positives and negatives.

Considering the situation that although point clouds contain much spatial information, it is inevitable to lose some in the process of downscaled feature maps with existing one-shot methods. 
He \textit{et al.} proposed a SA-SSD network exploiting ﬁne-grained architecture information to improve localization accuracy \cite{DBLP:conf/cvpr/HeZH0Z20}. 
What makes it speciﬁcal is that they ﬁrst transfer point cloud data to a tensor and put them to a backbone model for the purpose of extracting multiple stages features. 
What’s more, they designed a supplementary network in the possession of point-level supervision, which can instruct those features to learn the structure of point clouds. 
It is surprising that their experimental results demonstrate that SA-SSD ranks the ﬁrst on the KITTI BEV detection benchmark on the Car class.

\begin{figure*}
\begin{center}
\includegraphics[width=0.95\linewidth]{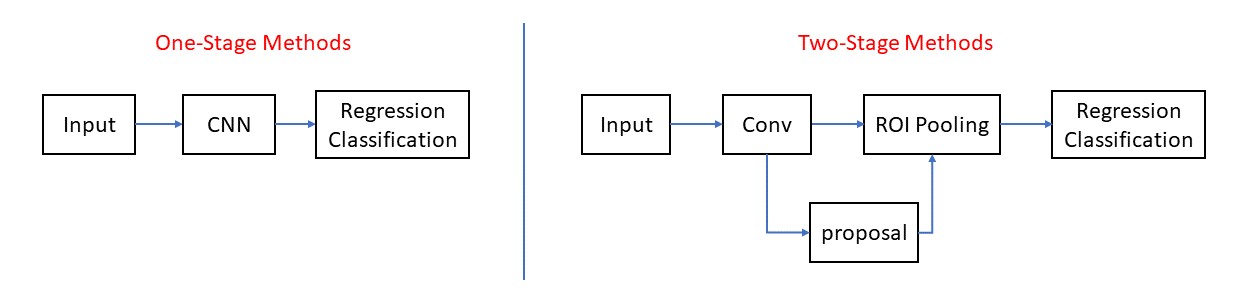}
\end{center}

   \caption{Different object detection processes for two-stage and one-stage methods.
}
\label{fig:difference}
\end{figure*}

\textbf{Point-based Methods.} These methods don’t transform the format of raw input, directly feeding the point cloud to the network. Yang \textit{et al.} did a pioneering work proposing a network named 3DSSD \cite{DBLP:conf/cvpr/YangS0J20}. This is the ﬁrst lightweight and efﬁcient point cloud-based 3D one-stage object detector. In this new model, Distance-FPS (D-FPS) and Feature-FPS (F-FPS) combined together are their proposed strategies for fusion sampling. In addition, they generated a detailed bounding box prediction network to get the best use of the representative points, exploiting a candidate generation layer (CG), an anchor-free regression head and a 3D centerness assignment strategy. At last, those experimental results demonstrate that 3DSSD outperforms the point-based method PointRCNN at a speed of 25 fps.

\textbf{Other Methods.} Also there are some other single-stage object detection methods which haven’t been divided into any types of method in this review. Meyer \textit{et al.} proposed LaserNet, which is an efﬁcient probabilistic 3D object detection model \cite{DBLP:conf/cvpr/MeyerLKVW19}. 
It is deserving to know that they use a small and dense range view data as input rather than Bird’s Eye View data. So their model is more efﬁcient. Further, to great knowledge, this is the ﬁrst method to obtain the uncertainty of detection in the manner of modelling the distribution of bounding box corners. 

Compared to two-stage methods, one-stage methods may don’t achieve such high detection accuracy. But one-stage methods have the ability to detect objects faster, which are suitable for real-time detection.

\section{Summary}

3D object detection is an enormous help for computers to understand scenes and is the key technique for a number of real-world applications such as autonomous driving. In this review, we list some typical state-of-the-art 3D object detection methods and categorize them into two-stage methods and one-stage methods. The former one needs to generate a series of proposals at ﬁrst and then predict of regress those extracted features. The one-stage method skips the procedure of proposal generation, directly predicting class probabilities and regressing bounding boxes. For the purpose of directly understanding how those two types of methods achieve object detection, Figure \ref{fig:difference} gives a simple description. We also state the advantages of point cloud data for object detection and list several acknowledged drawbacks of point clouds. At present, current popular methods try to introduce different types of input such as LIDAR points and camera data. The images provide us with more dense information but with the loss of 3D spatial information. LiDAR point cloud is suitable for 3D object detection with its geometric position information and depth information. The sparsity and the irregularity of point clouds urge people to investigate novel methods to leverage the advantages of both images and LiDAR-based data.

According to the analysis of those various kinds of existing methods, the following problems require further research:

\begin{itemize}
    \item First of all, owing to the regular representation of data, those mature 2D image process networks
can be applied to projection-based techniques and discretization-based techniques greatly. Nevertheless, it is inevitable to lose some useful information during the process of projecting 3D data into 2D format, which is a great limitation for projection-based methods. 
For discretization-based methods the exponentially increasing computation and huge memory costs caused by the increase of the resolution maintain the major bottleneck. 
Taking the above problems into consideration, building a sparse convolutional layer based on indexing architectures may be a feasible solution to those questions and it deserves further research.

    \item At present, point cloud-based models are popular methods that people pay the most attention to. However, point representation usually lacks clear neighbouring information because of the sparsity and irregularity of point clouds. A great many existing point cloud-based methods use expensive nearest neighbour searching techniques such as KNN applied in \cite{DBLP:conf/nips/LiBSWDC18}. The weak efﬁciency of these methods calls for more efﬁcient methods. A recent point-voxel combined representation method \cite{DBLP:conf/nips/LiuTLH19} can be a possible direction for further study.

    \item Most of the existing 3D point cloud object detection methods work on small scale of point clouds. However, those point cloud data obtained by LiDARs are extremely immense and large-scale because the process of data acquisition is continuous. As a result, there is a real thirst for further investigation to solve the problem of those large-scale point clouds.

    \item A large number of researchers \cite{DBLP:journals/corr/abs-1910-08287,DBLP:conf/iccv/LiuYB19} have begun to learn spatio-temporal information from dynamic point clouds. The spatio-temporal information is expected to help improve the performance of many later assignments such as 3D object segmentation, object recognition and completion.
    
\end{itemize}

{\small
\bibliographystyle{IEEEtran}
\bibliography{egbib}
}

\end{document}